\documentclass{article}

\usepackage[final, nonatbib]{neurips_2020}

\usepackage[utf8]{inputenc} 
\usepackage[T1]{fontenc}    
\usepackage{hyperref}       
\usepackage{url}            
\usepackage{booktabs}       
\usepackage{amsfonts}       
\usepackage{nicefrac}       
\usepackage{microtype}      

\usepackage{amsmath, amssymb}
\usepackage{graphicx}
\usepackage{subcaption}
\usepackage{xcolor}
\usepackage{bm}
\usepackage{enumitem}
\usepackage{subcaption}

\newcommand{\ie}{i.e.~}
\newcommand{\eg}{e.g.~}
\newcommand{\wrt}{w.r.t.~}

\renewcommand{\vec}[1]{\bm{#1}}

\newcommand{\denseem}{$\mathcal{H}^{{\rm EM}}_{{\rm dense}}$}
\newcommand{\densedirect}{$\mathcal{H}^{{\rm direct}}_{{\rm dense}}$}
\newcommand{\stand}{$\mathcal{H}_{{\rm stand}}$}
\newcommand{\standfair}{$\mathcal{H}^{{\rm fair}}_{{\rm stand}}$}

\title{DenseHMM: Learning Hidden Markov Models by Learning Dense Representations}

\author{
    {\small{Joachim Sicking$^{1,2}$, Maximilian Pintz$^{1,3}$, Maram Akila$^{1}$, Tim Wirtz$^{1,2}$}}\\
    $^1$ \small{Fraunhofer IAIS, Sankt Augustin, Germany}\\
    $^2$ \small{Fraunhofer Center for Machine Learning, Sankt Augustin, Germany}\\
    $^3$ \small{University of Bonn, Bonn, Germany}\\
    \scriptsize{\texttt{\{joachim.sicking, maximilian.alexander.pintz, maram.akila, tim.wirtz\}@iais.fraunhofer.de}}
    }
 
\begin{document}

\maketitle
\begin{abstract}
We propose \textit{DenseHMM} – a modification of Hidden Markov Models (HMMs) that allows to learn dense representations of both the hidden states and the observables. Compared to the standard HMM, transition probabilities are not atomic but composed of these representations via kernelization. Our approach enables constraint-free and gradient-based optimization. We propose two optimization schemes that make use of this: a modification of the Baum-Welch algorithm and a direct co-occurrence optimization. The latter one is highly scalable and comes empirically without loss of performance compared to standard HMMs. We show that the non-linearity of the kernelization is crucial for the expressiveness of the representations. The properties of the DenseHMM like learned co-occurrences and log-likelihoods are studied empirically on synthetic and biomedical datasets.
\end{abstract}

\section{Introduction}\label{sec:intro}
Hidden Markov Models \cite{rabiner1986introduction} have been a state-of-the-art approach for modelling sequential data for more than three decades \cite{hinton2012deep}. Their success story is proven by a large number of applications ranging from natural language modelling \cite{chen1999empirical} over financial services \cite{bhusari2016study} to robotics \cite{fu2016dynamic}. While still being used frequently, many more recent approaches are based on neural networks instead, like feed-forward neural networks~\cite{schmidhuber2015deep}, recurrent neural networks~\cite{hochreiter1997long} or spiked neural networks~\cite{tavanaei2019deep}. However, the recent breakthroughs in the field of neural networks~\cite{lecun2015deep,schmidhuber2015deep,goodfellow2016deep,minar2018recent,deng2013recent,paul2015review,wu2020recent} are accompanied by an almost equally big lack of their theoretical understanding. In contrast, HMMs come with a broad theoretical understanding, for instance of the parameter estimation~\cite{yang2017statistical}, convergence~\cite{wu1983convergence,dempster1977maximum}, consistency~\cite{leroux1992maximum} and short-term prediction performance~\cite{sharan2018prediction}, despite of their non-convex optimization landscape. 

Different from HMMs, (neural) representation learning became more prominent only recently~\cite{mikolov2013distributed, pennington2014glove}. From the very first day following their release, approaches like word2vec or Glove~\cite{mikolov2013distributed, pennington2014glove, mikolov2013efficient, le2014distributed} that yield dense representations for state sequences, have significantly emphasized the value of pre-trained representations of discrete sequential data for downstream tasks~\cite{kim2014convolutional,surveywordemb}. Since then those found application not only in language modeling~\cite{mikolov2013linguistic,zhang2016neural,li2018word,almeida2019word} but also in biology~\cite{asgari2015continuous,zou2019gene2vec}, graph analysis \cite{perozzi2014deepwalk, grover2016node2vec} and even banking~\cite{baldassini2018client2vec}. Similar approaches have received an overwhelming attention and became part of the respective state-of-the-art approaches.

Ever since, the quality of representation models increased steadily, driven especially by the natural language community. Recently, so-called transformer networks \cite{vaswani2017attention} were put forward, complex deep architectures that leverage attention mechanisms \cite{bahdanau2014neural, kim2017structured}. Their complexity and tremendously large amounts of compute and training data lead again to remarkable improvements on a multitude of natural language processing (NLP) tasks~\cite{devlin2018bert}.  

These developments are particularly driven by the question on \textit{how to optimally embed discrete sequences into a continuous space}. However, many existing approaches identify optimality solely with performance and put less emphasize on aspects like conceptual simplicity and theoretical soundness. Intensified discussions on \textit{well-understood} and therefore \textit{trustworthy} machine learning~\cite{saltzer1975protection,dwork2012fairness,amodei2016concrete,gu2017badnets,varshney2019trustworthy,toreini2020relationship,brundage2020toward} indicate, however, that these latter aspects become more and more crucial or even mandatory for real-world learning systems. This holds especially true when representing biological structures and systems as derived insights may be used in downstream applications, \eg of medical nature where physical harm can occur.

\begin{figure}
    \centering
    \includegraphics[width=0.95\textwidth]{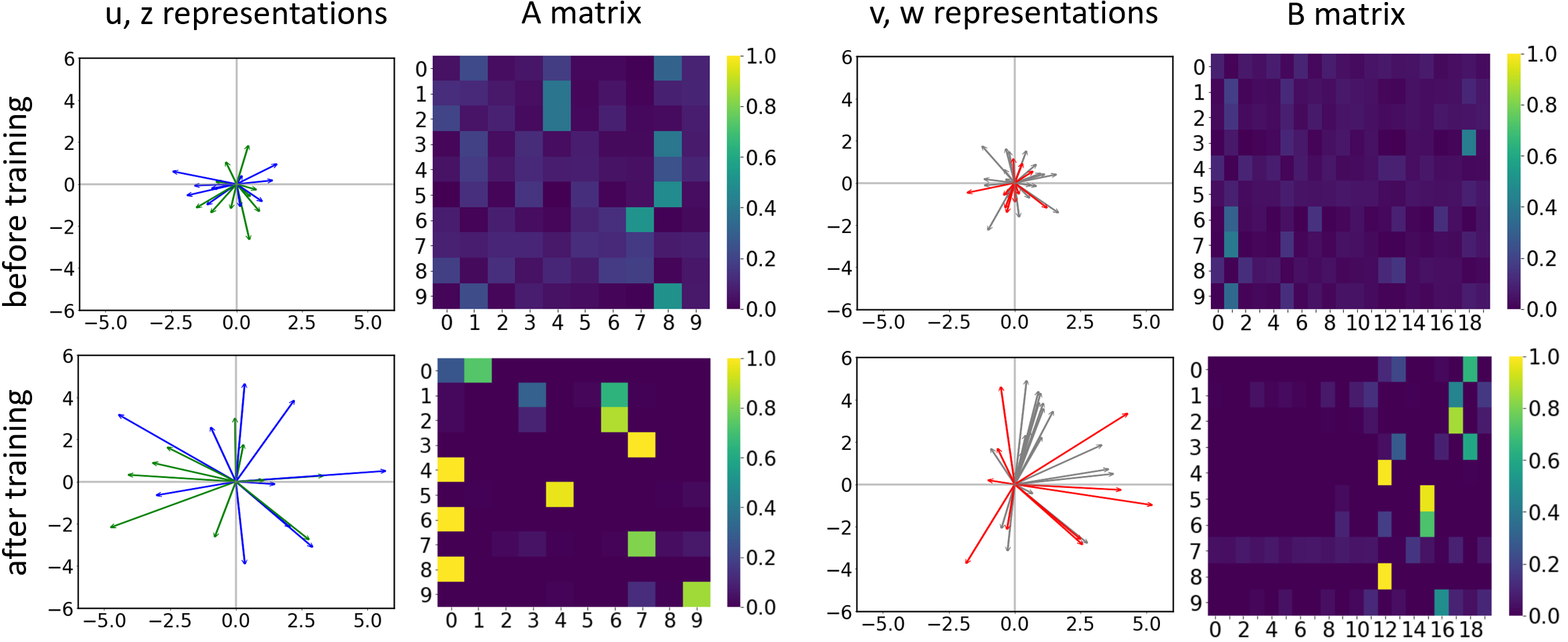}
    \caption{Exemplary DenseHMM to visualize the inner workings of our approach. All model components are shown before (top row) and after training (bottom row). The transition matrices $\bm{A}$ (second column) and $\bm{B}$ (fourth column) are learned by learning dense representations (first and third column). All representations are initialized by a standard Gaussian.}
    \label{fig:exemplary_dense_hmm}
\end{figure}

In light of this, we propose \textit{DenseHMM} – a modification of Hidden Markov Models that allows to learn dense representations of both the hidden states and the observables \mbox{(Figure \ref{fig:exemplary_dense_hmm})}. Compared to the standard HMM, transition probabilities are not atomic but composed of these representations. Concretely, we contribute
\begin{itemize}[itemsep=0.1cm,topsep=0pt,partopsep=0ex]
    \item a parameter-efficient, non-linear matrix factorization for HMMs,
    \item two competitive approaches to optimize the resulting DenseHMM 
    \item and an empirical study of its performance and properties on diverse datasets.
\end{itemize}
The rest of the work is organized as follows: first, we present related work on HMM parameter learning and matrix-factorization approaches in section \ref{sec:related_work}. Next, DenseHMM and its optimization schemes are introduced in section \ref{sec:structure-and-opt}. We study the effect of its softmax non-linearity and conduct empirical analyses and comparisons with standard HMMs in sections \ref{sec:theo_considerations} and  \ref{sec:emp_evaluation}, respectively. A discussion in section \ref{sec:discussion} concludes our paper. 

\section{Related work}\label{sec:related_work}

HMMs are generative models with Markov properties for sequences of either discrete or continuous observation symbols~\cite{rabiner1989tutorial}. They assume a (small) number of non-observable (hidden) states that drive the dynamics of the generated sequences. If domain expertise allows to interpret these drivers, HMMs can be fully understood. The simple discrete latent space as well as the explicit focus on short-term dependencies, are two properties that distinguish HMMs from sequence-modelling neural networks like long short-term memory networks \cite{hochreiter1997long} and temporal convolutional networks \cite{bai2018empirical}. More recent latent variable models that keep the discrete structure of the latent space make use of, e.g., Indian buffet processes \cite{griffiths2011indian}. These allow to dynamically adapt the dimension of the latent space dependent on data complexity and thus afford more flexible modelling. While we stay in the HMM model class, we argue that our approach allows to extend or reduce the latent space in a more principled way compared to standard HMMs.

Various approaches exist to learn the parameters of hidden Markov models: A classical one is the Baum-Welch algorithm \cite{rabiner1989tutorial} that handles the complexity of the joint likelihood of hidden states and observables by introducing an iterative two-step procedure that makes use of the forward-backward algorithm \cite{rabiner1986introduction}. Another algorithm for (local) likelihood maximization is \cite{baldi1994smooth}. The authors of \cite{pmlr-v80-huang18c} study HMM learning on observation co-occurrences instead of observation sequences. Based on moments, \ie co- and triple-occurrences, bounds on the empirical probabilities can be derived via spectral decomposition \cite{methodOfMoments}. Approaches from Bayesian data analysis comprise Markov chain Monte Carlo (MCMC) and variational \mbox{inference (VI)}. While MCMC can provide more stable and better results \cite{rybert2016}, it traditionally suffers from poor scalability. A more scalable stochastic-gradient MCMC algorithm that tackles mini-batching of sequentially dependent data is \cite{ma2017stochastic}. The same authors propose a stochastic VI (SVI) algorithm \cite{foti2014stochastic} that shares some technical details with \cite{ma2017stochastic}. SVI for hierarchical Dirichlet process (HDP) HMMs is considered in \cite{zhang2016stochastic}. For our DenseHMM, we adapt two non-Bayesian procedures: the Baum-Welch algorithm \cite{rabiner1989tutorial} and direct co-occurrence optimization \cite{pmlr-v80-huang18c}. The latter we optimize, solely for convenience, using a deep learning framework.\footnote{See appendix \ref{appendix:datapreproc} for details.} In \cite{NeuralHMMs} this idea was carried further, allowing different modifications to the original HMM context.

Non-negative matrix factorization (NMF, \cite{lee1999learning}) splits a matrix into a pair of low-rank matrices with solely positive components. NMF for HMM learning is used, e.g., in \cite{lakshminarayanan2010non,cybenko2011learning}. In contrast, we combine matrix factorization with a non-linear kernel function to ensure non-negativity and normalization of the HMM transition matrices. Further foci of recent work on HMMs are identifiability \cite{pmlr-v80-huang18c}, \ie uniqueness guarantees for an obtained model, and optimized priors over transition \mbox{distributions \cite{qiao2015diversified}}.

\section{Structure and optimization of the DenseHMM}\label{sec:structure-and-opt}

\begin{figure}
    \centering
    \begin{minipage}{.475\textwidth}
    \includegraphics[width=\linewidth]{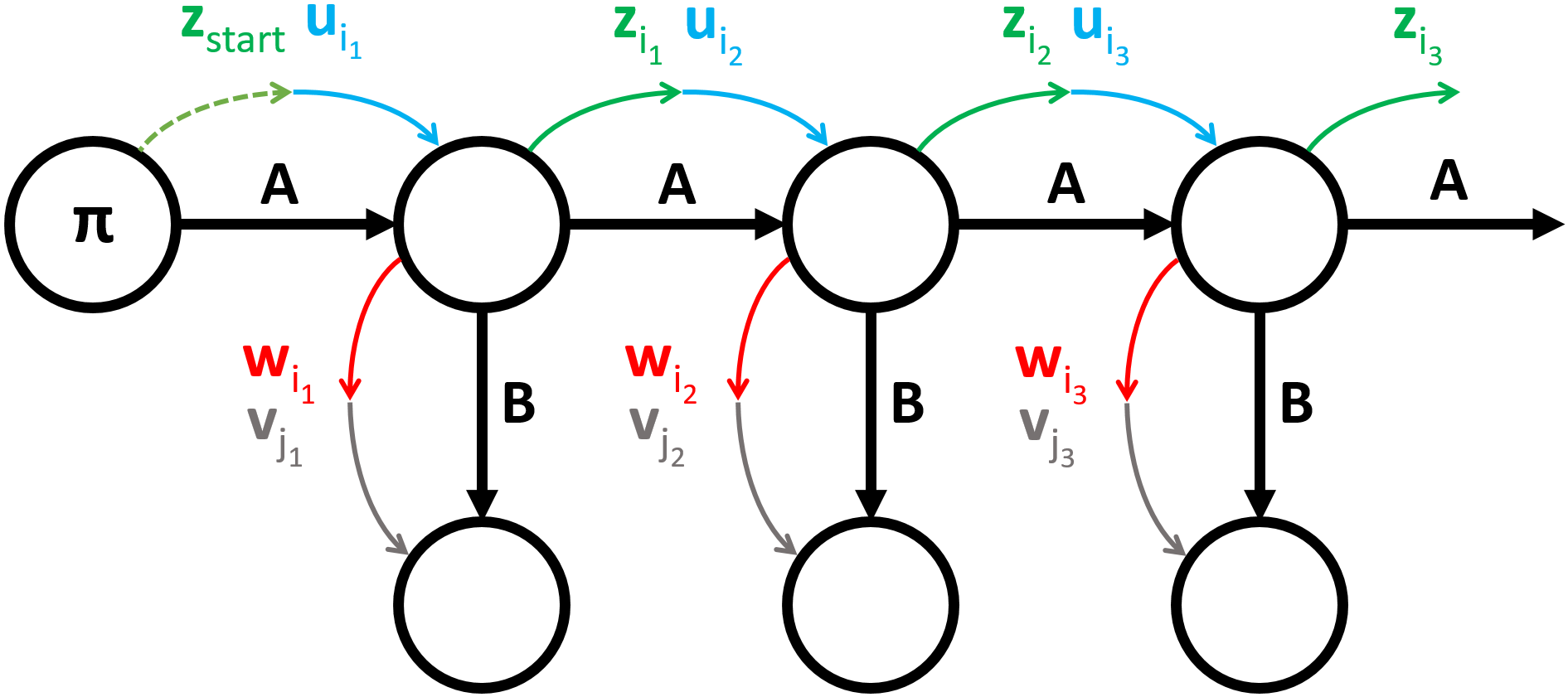}
    \captionof{figure}{Structure of the DenseHMM. The HMM parameters $\mathbf{A}$,$\textbf{B}$ and $\mathbf{\pi}$ are composed of vector representations such that $\mathbf{A} = \mathbf{A}(\mathbf{U}$,\,$\mathbf{Z})$, $\mathbf{B}=\mathbf{B}(\mathbf{V}$,$\mathbf{W})$ and $\mathbf{\pi}=\mathbf{\pi}(\mathbf{U},\mathbf{z}_{\rm start})$.}
    \label{fig:structure_densehmm}
    \end{minipage}%
    \begin{minipage}{.05\textwidth}
    ~
    \end{minipage}%
    \begin{minipage}{.475\textwidth}
    \includegraphics[width=\linewidth]{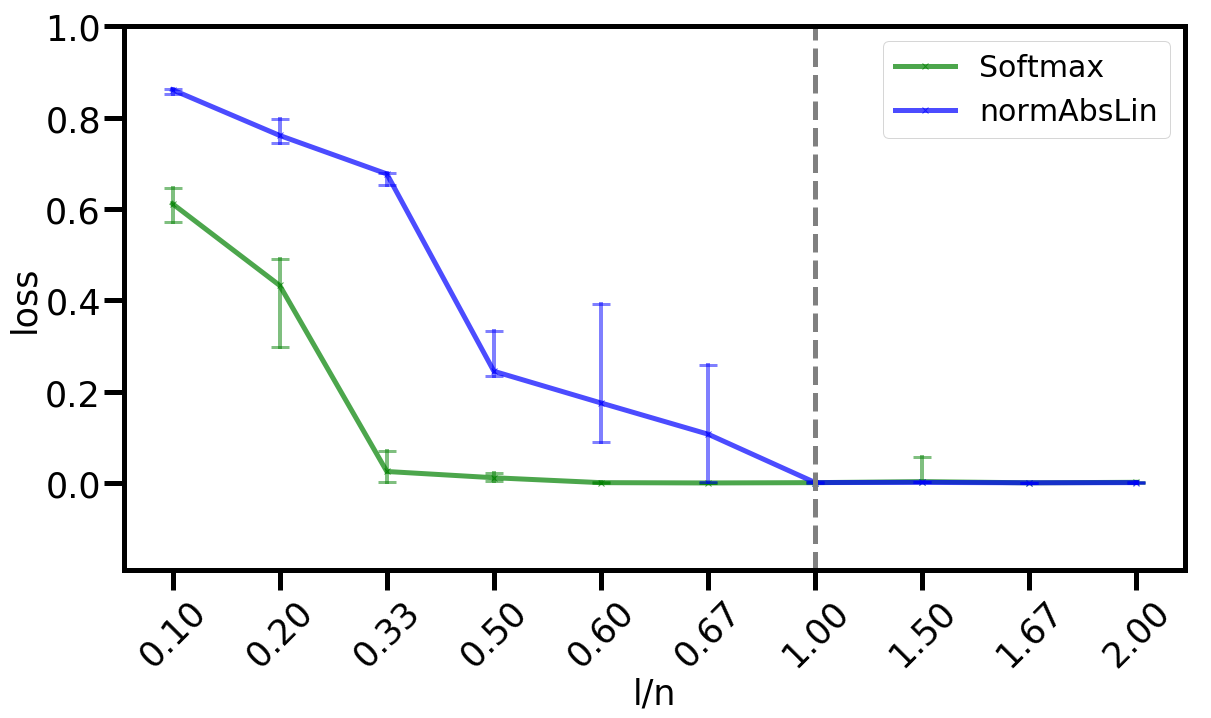}
    \captionof{figure}{Approximation quality of non-linear matrix factorizations. The optimization errors (median, 25/75 percentile) of softmax (green) and normAbsLin (blue) matrices are shown over the ratio of representation length $l$ and matrix size $n$. The vertical line indicates $l = n$.}
    \label{fig:nonlinear_matrix_approx}
    \end{minipage}
\end{figure}

A HMM is defined by two time-discrete stochastic processes: $\{X_t\}_{t\in\mathbb{N}}$ is a Markov chain over hidden states $S = \{s_i\}_{i=1}^n$ and $\{Y_t\}_{t\in\mathbb{N}}$ is a process over observable states $O = \{o_i\}_{i=1}^m$. The central assumption of HMMs is that the probability to observe $Y_t=y_t$ depends only on the current state of the hidden process and the probability to find $X_t=x_t$ only on the the previous state of the hidden process, $X_{t-1}=x_{t-1}$ for all $t \in \mathbb{N}$. We denote the state-transition matrix as $\bm{A} \in \mathbb{R}^{n\times n}$ with \mbox{$a_{ij} = P(X_t = s_j \mid X_{t-1} = s_i)$}, the emission matrix as $\bm{B} \in \mathbb{R}^{n\times m}$ with $b_{ij} = P(Y_t = o_j \mid X_t = s_i)$ and the initial state distribution as $\bm{\pi} \in \mathbb{R}^n$ with $\pi_i = P(X_1 = s_i)$. A HMM is fully parametrized by $\bm{\lambda} = (\bm{A}, \bm{B}, \bm{\pi})$.

HMMs can be seen as extensions of Markov chains (MCs) which are in turn closely related to word2vec embeddings. Let us elaborate on this: A MC has no hidden states and is defined by just one process $\{X_t\}_{t\in\mathbb{N}}$ over observables. 
The transition dynamics of the states of the MC is described by a transition matrix $\bm{A}$ and an initial distribution $\bm{\pi}$. Being in a given state $s_I$, the MC models conditional probabilities of the form $p(s_i \mid s_I)$. 
MCs are structurally similar to the approaches that learn word2vec representations, \ie continuous bag of words and skip-gram \cite{mikolov2013distributed}. Both models learn transitions between the words of a text corpus. Each word $w_i$ of the vocabulary is represented by a learned dense vector $\vec{u}_i$. The transition probabilities between words are recovered from the scalar products of these vectors:  
\begin{align}
    p(w_j \mid w_i) = \frac{\exp(\vec{u}_i\cdot \vec{v}_j)}{\sum_k \exp(\vec{u}_i\cdot \vec{v}_k)} \propto \exp(\vec{u}_i\cdot \vec{v}_j). \label{eq:w2vfac}
\end{align}
The learned word2vec representation are low-dimensional and context-based, \ie they contain semantic information. This is in contrast to the trivial and high-dimensional one-hot (or bag-of-word) encodings.

Here we transfer the non-linear factorization approach of word2vec (eq. \ref{eq:w2vfac}) to HMMs. This is done by composing $\bm{A}, \bm{B}$ and $\bm{\pi}$ of dense vector representations such that
\begin{subequations}
    \begin{alignat}{3}
        a_{ij} & = a_{ij}(\mathbf{U},\mathbf{Z}) &&=\ \frac{\exp(\vec{u}_{j} \cdot \vec{z}_{i})}{\sum\nolimits_{k \in [n]} \exp(\vec{u}_{k} \cdot \vec{z}_{i})}   &\qquad& \text{for $i,j \in [n]$,} \label{eq:atransf} \\
        b_{ij} & = b_{ij}(\mathbf{V},\mathbf{W}) &&=\ \frac{\exp(\vec{v}_{j} \cdot \vec{w}_{i})}{\sum\nolimits_{k \in [m]} \exp(\vec{v}_{k} \cdot \vec{w}_{i})} &\qquad& \text{for $i \in [n], j \in [m]$,} \label{eq:btranf} \\
        \pi_i & = \pi_i(\mathbf{U},\mathbf{z}_{\rm start}) &&=\ \frac{\exp(\vec{u}_{i} \cdot \vec{z}_{\rm{start}})}{\sum\nolimits_{k \in [n]} \exp(\vec{u}_{k} \cdot \vec{z}_{\rm{start}})} &\qquad& \text{for $i \in [n]$.} \label{eq:pitranf}
    \end{alignat}
\end{subequations}
Let us motivate this transformation (Fig. \ref{fig:structure_densehmm}) piece by piece: each representation vector corresponds to either a hidden state ($\bm{u}_i$,$\bm{w}_i$,$\bm{z}_i$) or an observation ($\bm{v}_i$). The vector $\bm{u}_i$ ($\bm{z}_i$) is the incoming (outgoing) representation of hidden state $i$ along the (hidden) Markov chain. $\bm{w}_i$ is the outgoing representation of hidden state $i$ towards  the observation symbols. These are described by the $\bm{v}_i$. All vectors are real-valued and of length $l$.
$\bm{A}$ and $\bm{B}$ each depend on two kinds of representations instead of only one to enable non-symmetric transition matrices. Additionally, to choose $\bm{A}$ independent of $\bm{B}$, as is typical for HMMs, we need $\bm{w}_i$ as a third hidden representation.
It is convenient to summarize all representation vectors of one kind in a matrix ($\bm{U}, \bm{V}, \bm{W}, \bm{Z}$).

A softmax kernel maps the scalar products of the representations onto the HMM parameters $\bm{A}, \bm{B}$ and $\bm{\pi}$. Softmax maps to the simplex and thus ensures $a_{ij}$, $b_{ij}$, $\pi_i$ to be in $[0,1]$ as well as row-wise normalization of $\bm{A}$, $\bm{B}$ and $\bm{\pi}$.

This non-linear kernelization enables constraint-free optimization which is a central property of our approach. We use this fact in two different ways: we derive a modified expectation-maximization (EM) scheme in section \ref{sec:em} and study an alternative to EM optimization that is based on co-occurrences in \mbox{section \ref{sec:cooc}}.

\subsection{EM optimization: a gradient-based M-step}\label{sec:em}
We briefly recapitulate the EM-based Baum-Welch algorithm~\cite{bishop2007} and adapt it to learn the proposed representations as part of the M-step: 

Given a sequence $\vec{o}$ of length $T \in\mathbb{N}$ over observations $O$, the Baum-Welch algorithm finds parameters $\lambda$ that (locally) maximize the observation likelihoods.
A latent distribution $Q$ over the hidden states $S$ is introduced such that the log-likelihood of the sequence decomposes as follows:
$\mathfrak{L}(Q,\bm{\lambda}) = \log P(\vec{o}, \bm{\lambda}) = \mathcal{L}(Q, \bm{\lambda}) + KL(Q || P(\cdot \mid \vec{o}, \bm{\lambda}))$. $KL(P||Q)$ denotes the Kullback-Leibler divergence from $Q$ to $P$ with $P, Q$ being probability distributions and \mbox{$\mathcal{L}(Q, \bm{\lambda}) = \sum_{\vec{x}\in S^T} Q(\vec{x}) \log\left[P(\vec{x}, \vec{o}; \bm{\lambda}) / Q(\vec{x})\right]$}.
Starting from an initial guess for $\bm{\lambda}$, the algorithm alternates between two sub-procedures, the E- and M-step: In the E-step, the forward-backward algorithm~\cite{bishop2007} is used to update
$Q = P(\cdot \mid \vec{o}; \bm{\lambda})$, which maximizes $\mathfrak{L}(Q, \bm{\lambda})$ for fixed $\bm{\lambda}$. The efficient computation of the conditional probabilities $\gamma_t(s, s') := P(X_{t-1}=s,X_t=s' \mid \vec{o})$ and $\gamma_t(s) := P(X_t = s \mid \vec{o})$ for $s, s' \in S$ is crucial for the E-step.
In the M-step, the latent distribution $Q$ is fixed and $\mathfrak{L}(Q, \bm{\lambda})$ is maximized \wrt $\bm{\lambda}$ under normalization constraints. 
As the Kullback-Leibler divergence $KL$ is set to zero in each E-step, the function to maximize in the M-step becomes
\begin{align*}
    \mathcal{L}(Q, \bm{\lambda})  = \sum_{\vec{x}\in S^T} Q(\vec{x}) \log\frac{P(\vec{x}, \vec{o}; \bm{\lambda})}{Q(\vec{x})} = \sum_{\vec{x} \in S^T} P(\vec{x} \mid \vec{o}; \bm{\lambda}^{\rm{old}}) \log \frac{P(\vec{x}, \vec{o}; \bm{\lambda})}{P(\vec{x} \mid \vec{o}; \bm{\lambda}^{\rm{old}})}
\end{align*}
with $\bm{\lambda}^{\rm{old}}$ being the parameter obtained in the previous M-step.
Applying the logarithm to $P(\vec{x},\vec{o}; \vec{\lambda})$, which has a product structure due to the Markov properties, splits the optimization objective into three summands. Each term depends on only one of the parameters $\bm{A}, \bm{B}, \bm{\pi}$:
\begin{align*}
    \bm{A}^*, \bm{B}^*, \bm{\pi}^* = \arg\max_{\bm{A},\bm{B},\bm{\pi}} \mathcal{L}(Q, \bm{\lambda}) = \arg\max_{\bm{A},\bm{B},\bm{\pi}} \mathcal{L}_1(Q, \bm{A}) + \mathcal{L}_2(Q, \bm{B}) + \mathcal{L}_3(Q, \bm{\pi}).
\end{align*}
Due to structural similarities between the three summands, we consider only $\mathcal{L}_1$ in the following. The treatment of $\mathcal{L}_2$ and $\mathcal{L}_3$ can be found in appendix \ref{appendix:full_lagrangians}. For $\mathcal{L}_1$, we have
\begin{align*}
    \mathcal{L}_1(Q, \bm{A}) = \sum_{\bm{i}\in [n]^T} P(s_{\bm{i}} \mid \bm{o}; \bm{\lambda}^{old}) \sum_{t=2}^T \log(a_{i_{t-1}, i_t})
\end{align*}
with multi-index $s_{\bm{i}} = s_{i_1}, ..., s_{i_T}$. The next step is to re-write $\mathcal{L}_1$ in terms of $\gamma_t$ and to use Lagrange multipliers to ensure normalization. This gives the following part $ \bar{\mathcal{L}}_1$ of the full Lagrangian $\bar{\mathcal{L}} = \bar{\mathcal{L}}_1 + \bar{\mathcal{L}}_2 + \bar{\mathcal{L}}_3$:
\begin{align*} 
\bar{\mathcal{L}}_1 := &\sum\limits_{i, j \in [n]} \sum_{t=2}^T \gamma_t(s_i, s_j) \log a_{ij} + \sum_{i \in [n]} \varphi_i \Big(1 - \sum_{j \in [n]} a_{ij}\Big)
\end{align*}
with Lagrange multipliers $\varphi_i$ for each $i \in [n]$ to ensure that $\bm{A}$ is a proper transition matrix. 

To optimize DenseHMM, we leave the E-step unchanged and modify the M-step by applying the parametrization of $\bm{\lambda}$, \ie equations (\ref{eq:atransf})-(\ref{eq:pitranf}), to the Lagrangian $\bar{\mathcal{L}}$. Please note that we can drop all normalization constraints as they are explicitly enforced by the softmax function.
This turns the original constrained optimization problem of the M-step into an unconstrained one, leading to a Lagrangian of the form $\bar{\mathcal{L}}^{\rm{dense}} = \bar{\mathcal{L}}^{\rm{dense}}_1 + \bar{\mathcal{L}}^{\rm{dense}}_2 + \bar{\mathcal{L}}^{\rm{dense}}_3$ with 
\begin{align*}
    \bar{\mathcal{L}}^{\rm{dense}}_1 = &\sum_{i, j \in [n]} \sum_{t=2}^T \gamma_t(s_i, s_j) \vec{u}_{j} \cdot \vec{z}_{i} - \sum_{i, j \in [n]}\sum_{t=2}^T \gamma_t(s_i, s_j) \log \sum_{ k \in [n]} \exp(\vec{u}_k \cdot \vec{z}_{i}).
\end{align*}
We optimize $\bar{\mathcal{L}}^{\rm{dense}}$ with gradient-decent procedures such as SGD \cite{bottou2010large} and Adam \cite{kingma2014adam}. 

\subsection{Direct optimization of observation co-occurrences: gradient-based and scalable}\label{sec:cooc}

Inspired by \cite{pmlr-v80-huang18c}, we investigate an alternative to the EM scheme: directly optimizing co-occurrence probabilities. The ground truth co-occurrences $\bm{\Omega}^{\rm gt}$ are obtained from training data $\vec{o}$ by calculating the relative frequencies of subsequent pairs $(o_i(t), o_j(t+1)) \in O^2$. If we know that $\vec{o}$ is generated by a HMM with a stationary hidden process, we can easily compute $\bm{\Omega}^{\rm gt}$ analytically as follows: we summarize all co-occurrence probabilities $\Omega_{ij} = P(Y_t\!=\!o_i, Y_{t+1}\!=\!o_j)$ for $i, j \in [m]$ in a co-occurrence matrix $\bm{\Omega} = \bm{B}^T \bm{\Theta} \bm{B}$ with $\Theta_{kl} = P(X_t\!=\!s_k, X_{t+1}\!=\!s_l)$ for $k, l \in [n]$. We can further write $\Theta_{kl} = P(X_{t+1}\!=\!s_l\!\mid\!X_t\!=\!s_k)\,P(X_t\!=\!s_k) = A_{kl}\,\pi_k$ for $i, j \in [n]$ under the assumption that $\vec{\pi}$ is the stationary distribution of $\bm{A}$, \ie $\pi_j = \sum_i A_{ij}\,\pi_i$ for all $i, j \in [n]$. Then, we obtain the co-occurrence probabilities
\begin{alignat}{2} \label{eq:coocs}
   \Omega_{ij} &= \sum_{k,l \in [n]} \pi_{k}\,b_{ki}\,a_{kl}\,b_{lj} &\qquad&\text{ for } i,j \in [m].
\end{alignat}
Parametrizing the matrices $\bm{A}$ and $\bm{B}$ according to eq. (\ref{eq:atransf},\ref{eq:btranf}) yields
\begin{align*}
    \Omega^{\rm{dense}}_{ij}(\bm{U}, \bm{V}, \bm{W}, \bm{Z}) = \sum_{k,l \in [n]} \pi_{k}\,b_{ki}(\mathbf{V},\mathbf{W})\, a_{kl}(\mathbf{U},\mathbf{Z})\,b_{lj}(\mathbf{V},\mathbf{W}) \ \ \text{ for } i,j \in [m].
\end{align*} 
Please note that $\bm{\pi}$ is not parametrized here. Following our stationarity demand it is chosen as the eigenvector $\mathbf{v}_{\lambda = 1}$ of $\bm{A}^T$. We minimize the squared distance between $\bm{\Omega}^{\rm dense}$ and $\bm{\Omega}^{\rm gt}$ \wrt the vector representations, \ie 
\begin{align*}
    {\arg\min}_{\bm{U}, \bm{V}, \bm{W}, \bm{Z}}\ || \bm{\Omega}^{\rm gt} - \bm{\Omega}^{\rm dense}(\bm{U}, \bm{V}, \bm{W}, \bm{Z}) ||^2_F\,,
\end{align*}
using gradient-decent procedures like SGD and Adam.

\section{Properties of the DenseHMM}\label{sec:theo_considerations}
To further motivate our approach, please note that a standard HMM with $n$ hidden states and $m$ observation symbols has $n^2 + n(m - 1) - 1$ degrees of freedom (DOFs), whereas a DenseHMM with representation length $l$ has $l(3n + m + 1)$ DOFs. Therefore, a low-dimensional representation length $l$ leads to DenseHMMs with less DOFs compared to a standard HMM for many values of $n$ and $m$. A linear factorization with representation length $l<n$ leads to rank $l$ for the matrices $\bm{A}$ and $\bm{B}$, whereas a non-linear factorization can yield more expressive full rank matrices. This effect of non-linearities may be best understood with a simple toy example: assume a 2x2 matrix with co-linear columns: $[[1,2],[2,4]]$. Applying a softmax column-wise leads to a matrix $\propto [[e,e^2],C[e^2,e^4]]$ with linearly independent columns. More general, the softmax rescales and rotates each column of $\bm{UZ}$ and $\bm{VW}$ differently and (except for special cases) thus increases the matrix rank to full rank. It is worth to mention that any other kernel $k:\mathbb{R}^l \times \mathbb{R}^l \rightarrow \mathbb{R}^+$ could be used instead of exp in the softmax to recover the HMM parameters from the representations, e.g. sigmoid, ReLU or RBFs. 

As non-linear matrix factorization is a central building block of our approach, we compare the approximation quality of softmax with an appropriate linear factorization in the following setup: we generate a Dirichlet-distributed ground truth matrix $\bm{A}_{\rm gt} \in \mathbb{R}^{n\times n}$ and approximate it (i) by $\tilde{\bm{A}} = \rm{softmax}(\mathbf{UZ})$ defined by $\rm{softmax}(\mathbf{UZ})_{ij} = \exp\left((\mathbf{UZ})_{ij}\right) / \sum_k \exp\left((\mathbf{UZ})_{ik}\right)$ and (ii) by a normalized absolute matrix product $\tilde{\bm{A}} = \rm{normAbsLin}(\mathbf{UZ})$ defined by \mbox{$\rm{normAbsLin}(\mathbf{UZ})_{ij} = |(\mathbf{UZ})_{ij}| / \sum_k |(\mathbf{UZ})_{ik}|$}. Note that we report the resulting error $||\tilde{\bm{A}} - \bm{A}_{\rm gt}||_F$ divided by $||\bm{A}_{\rm gt}||_F$ to get comparable losses independent of the size of $\bm{A}_{\rm gt}$. These optimizations are performed for matrix sizes $n = {3,5,10}$, several representation lengths $l$ and $10$ different $\bm{A}_{\rm gt}$ for each $(n,l)$ pair. Table \ref{expd} in appendix A provides all considered $(n,l)$ pairs and detailed results. Fig. \ref{fig:nonlinear_matrix_approx} shows that the softmax non-linearity yields closer approximations of $\bm{A}_{\rm gt}$ compared to normAbsLin. Moreover, we observe on a qualitative level significantly faster convergence for softmax as $l$ increases. For softmax, vector lengths $l \approx n/3$ suffice to closely fit $\bm{A}_{\rm gt}$ while the piece-wise linear normAbsLin requires $l = n$. This result is in accordance with our remarks above.

\section{Empirical evaluation}\label{sec:emp_evaluation}

We investigate the outlined optimization schemes \wrt obtained model quality and behaviors. We compare the following types of models:
\begin{itemize}[itemsep=0.1cm,topsep=0pt,partopsep=0ex, leftmargin=2cm]
    \item[$\mathcal{H}^{{\rm EM}}_{{\rm dense}}$:] a DenseHMM optimized with the EM optimization scheme (section \ref{sec:em}),
    \item[$\mathcal{H}^{{\rm direct}}_{{\rm dense}}$:] a DenseHMM optimized with direct optimization of co-occurrences (section \ref{sec:cooc}),
    \item[$\mathcal{H}_{{\rm stand}}$:] a standard HMM optimized with the Baum-Welch algorithm \cite{rabiner1989tutorial}.
\end{itemize}
These models all have the same number of hidden states $n$ and observation symbols $m$. If a standard HMM and a DenseHMM use the same $n$ and $m$, one of the models may have less DOFs than the other (cp. section \ref{sec:theo_considerations}). Therefore, we also consider the model $\mathcal{H}^{{\rm fair}}_{{\rm stand}}$, which is a standard HMM with a similar amount of DOFs compared to a given $\mathcal{H}_{\rm{dense}}$ model. We denote the number of hidden states in $\mathcal{H}_{{\rm stand}}^{{\rm fair}}$ as $n_{{\rm fair}}$, which is the positive solution of $n_{{\rm fair}}^2 + n_{{\rm fair}}(m - 1) - 1 = l(3n + m + 1)$ rounded to the nearest integer. Note that $n_{{\rm fair}}$ can get significantly larger than $n$ for $l > n$ and therefore \standfair~ is expected to outperform the other models in these cases.

We use two standard measures to assess model quality: the co-occurrence mean absolute \mbox{deviation (MAD)} and the normalized negative log-likelihood (NLL). The MAD between two co-occurrence matrices $\bm{\Omega}^{\rm gt}$ and ${\bm \Omega}^{\rm model}$ is defined as $1/m^2 \sum_{i, j \in [m]} |\Omega_{ij}^{{\rm model}} - \Omega^{{\rm gt}}_{ij}|$. We compute both $\bm{\Omega}^{\rm gt}$ and $\bm{\Omega}^{\rm model}$ based on sufficiently long sampled sequences (more details in appendix \ref{appendix:datapreproc}). 
In the case of synthetically generated ground truth sequences, we compute $\bm{\Omega}^{\rm gt}$ analytically instead. In addition, we take a look at the negative log-likelihood of the ground truth test sequences $\{\vec{o}^{{\rm test}}_i\}$ under the model, \ie ${\rm NLL} = - \sum_i \log P(\vec{o}^{{\rm test}}_i; \vec{\lambda})$. 
We conduct experiments with $n \in \{3, 5, 10\}$ and different representation lengths $l$ for each $n$. For each $(n, l)$ combination, we run $10$ experiments with different train-test splits. We evaluate the median and 25/75 percentiles of the co-occurrence MADs and the normalized NLLs for each of the four models (see appendix \ref{appendix:datapreproc} for details).

In the following, we consider synthetically generated data as well as two real-world datasets: amino acid sequences from the RCSB PDB dataset \cite{protein} and part-of-speech tag sequences of biomedical text \cite{medpost}, referred to as the MedPost dataset.

\begin{figure}
   \begin{subfigure}{.5\textwidth}
   \centering
   \includegraphics[height=0.198\textheight]{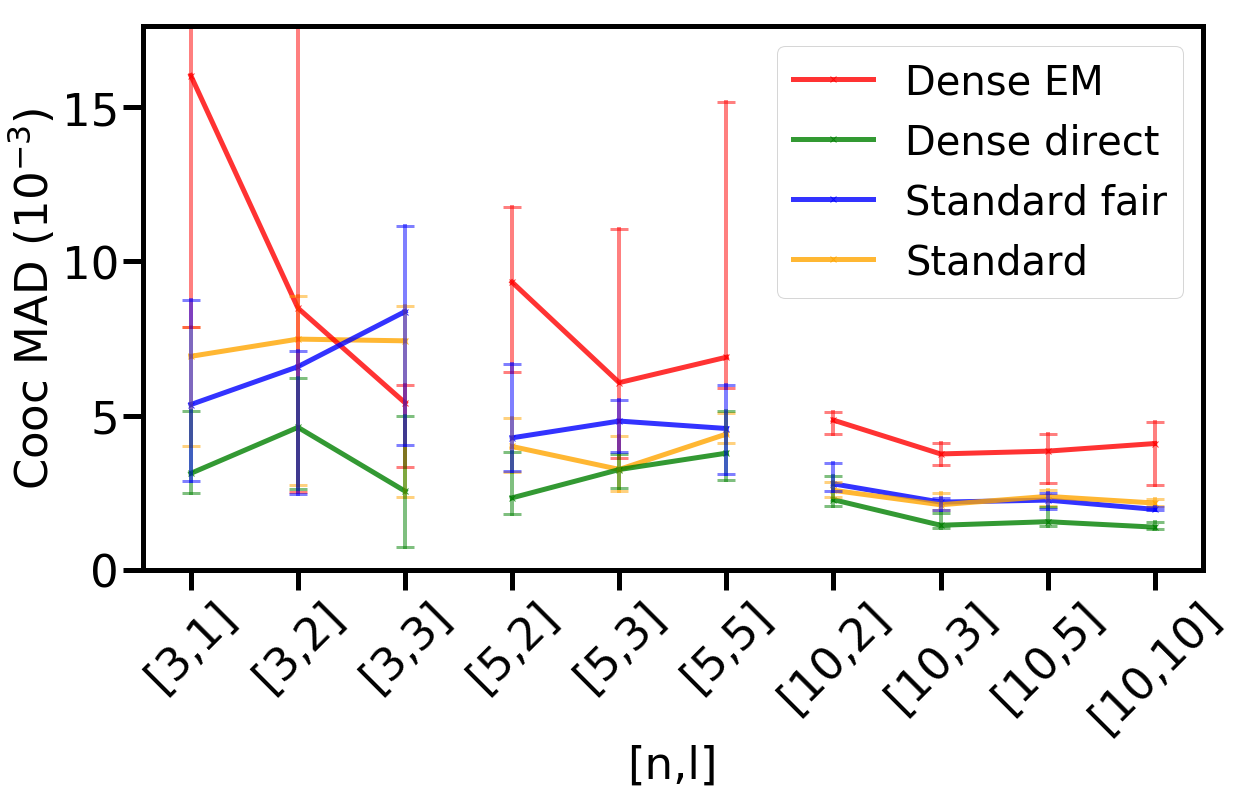}
   \end{subfigure}
    \begin{subfigure}{.5\textwidth}
    \centering
    \includegraphics[height=0.198\textheight]{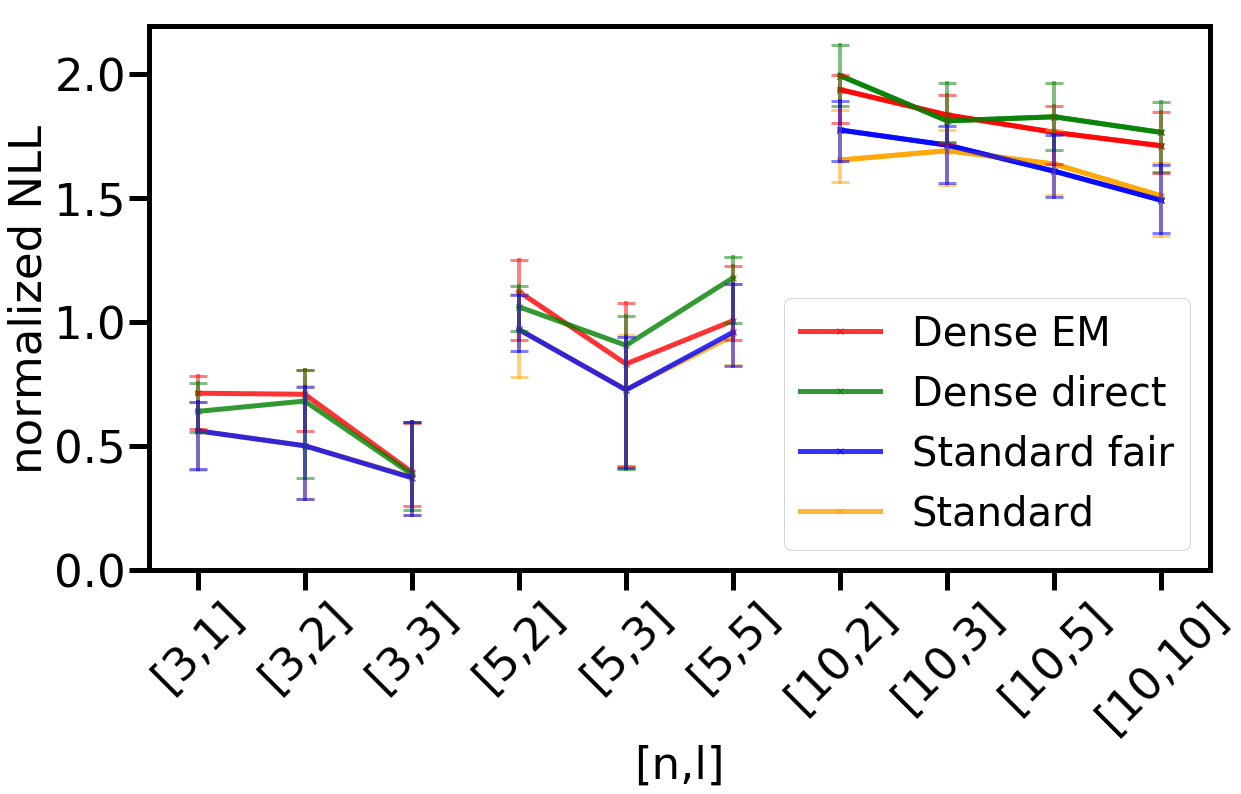}
    \end{subfigure}
    \caption{Co-occurrence mean absolute deviation (left) and normalized negative log-likelihood (right) of the models \denseem~, \densedirect~, \standfair~, \stand~ on synthetically generated sequences evaluated for multiple combinations of $n$ and $l$. }
    \label{fig:synth}
\end{figure}

\begin{figure}
   \begin{subfigure}{.5\textwidth}
   \centering
   \includegraphics[height=0.198\textheight]{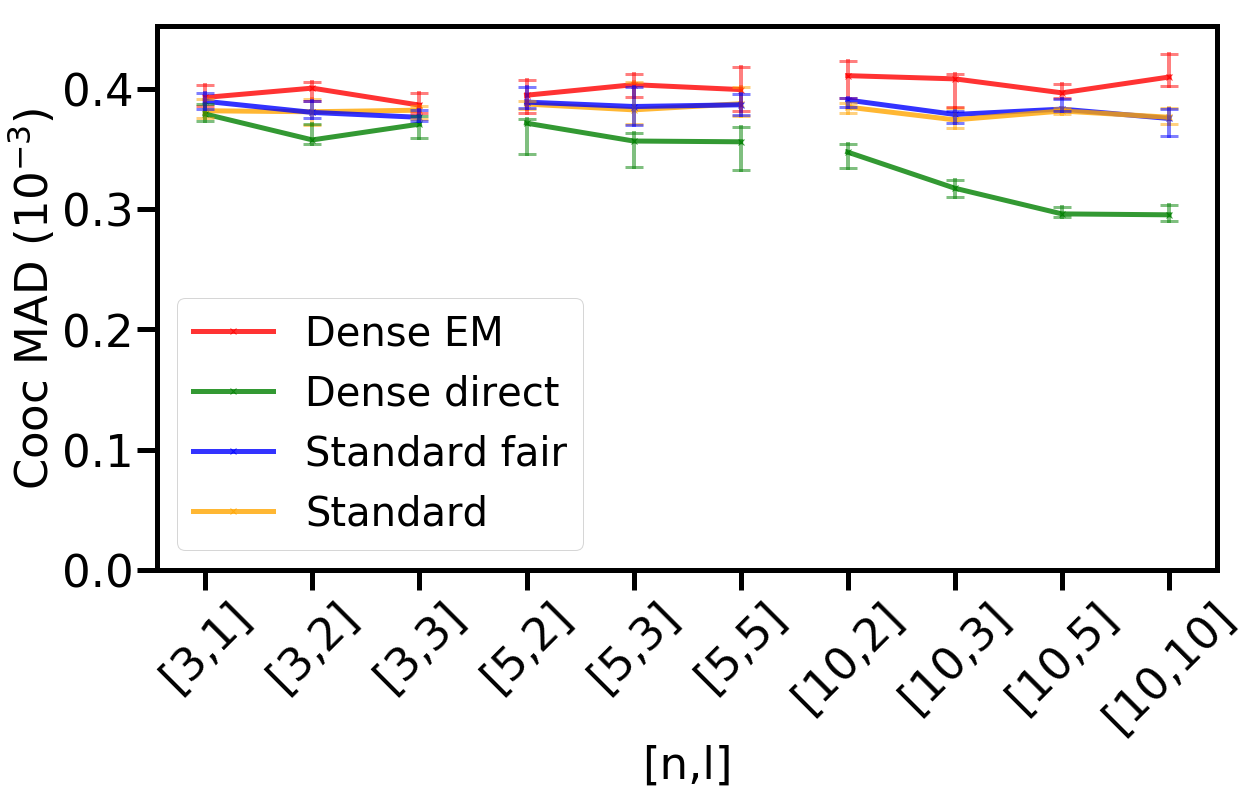}
   \end{subfigure}
    \begin{subfigure}{.5\textwidth}
    \centering
    \includegraphics[height=0.198\textheight]{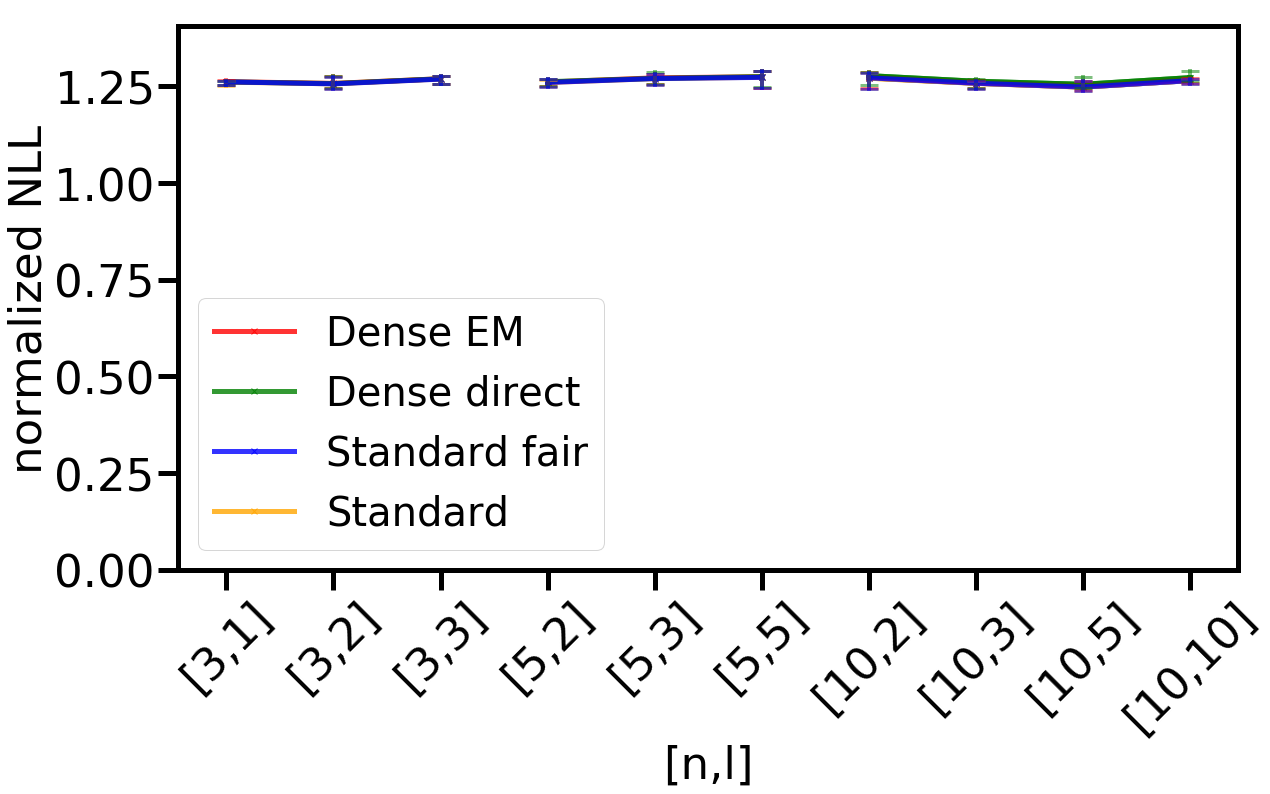}
    \end{subfigure}
    \caption{Co-occurrence mean absolute deviation (left) and normalized negative log-likelihood (right) of the models \denseem~, \densedirect~, \standfair~, \stand~ on amino acid sequences evaluated for multiple combinations of $n$ and $l$.}
    \label{fig:protein}
\end{figure}

\paragraph{Synthetic sequences} We sample training and test ground truth sequences from a standard HMM $\mathcal{H}_{\rm syn}$ that is constructed as follows: Each row of the transition matrices $\bm{A}$ and $\bm{B}$ is drawn from a Dirichlet distribution $Dir(\vec{\alpha})$, where all entries in $\vec{\alpha}$ are set to a fixed value $\alpha = 0.1$. The initial state distribution $\bm{\pi}$ is set to the normalized eigenvector ${\bm v}_{\lambda = 1}$ of ${\bm A}^T$. This renders $\mathcal{H}_{\rm syn}$ stationary and allows a simple analytical calculation of $\Omega^{{\rm\bm gt}}$ according to eq. \ref{eq:coocs}. For both training and testing, we sample $10$ sequences, each of length $200$ with $m = n$ emission symbols. Figure \ref{fig:synth} left shows our evaluation \wrt co-occurrence MADs. Note that the performance of \stand~ changes slightly with $l$ for fixed $n$ as training is performed on different sequences for every $(n, l)$ pair. This is because the sequences are re-drawn from $\mathcal{H}_{\rm syn}$ for every experiment. The standard HMMs and \densedirect~ perform similarly, with \densedirect~ performing slightly better throughout the experiments and especially for $n = 3$. \denseem~ shows a higher MAD than the other models. The good performance of \densedirect~ may be explained by the fact that it optimizes a function similar to co-occurrence MADs, whereas the other models aim to optimize negative log-likelihoods. The results in Figure \ref{fig:synth} right show that the DenseHMMs reach comparable NLLs, although the standard HMMs perform slightly better in this metric.
\paragraph{Proteins} The RCSB PDB dataset \cite{protein} consists of $512{,}145$ amino acid sequences from which we only take the first $1{,}024$. After applying preprocessing (described in further detail in appendix \ref{appendix:datapreproc}), we randomly shuffle the sequences and split train and test data 50:50 for each experiment. Figure \ref{fig:protein} left shows the results of our evaluation \wrt co-occurrence MADs. \denseem~ performs slightly worse than both \stand~ and \standfair. We observe that \densedirect~ yields the best results. While the co-occurrence MADs of \denseem, \stand~ and \standfair~ stay roughly constant throughout different experiments, \densedirect~ can utilize larger $n$ and $l$ to further decrease co-occurrence MADs. As can be seen in Figure \ref{fig:protein} right, all models achieve almost identical normalized NLLs throughout the experiments. The results suggest that model size has only a minor impact on normalized NLL performance for the protein dataset.

\begin{figure}
   \begin{subfigure}{.5\textwidth}
   \centering
   \includegraphics[height=0.198\textheight]{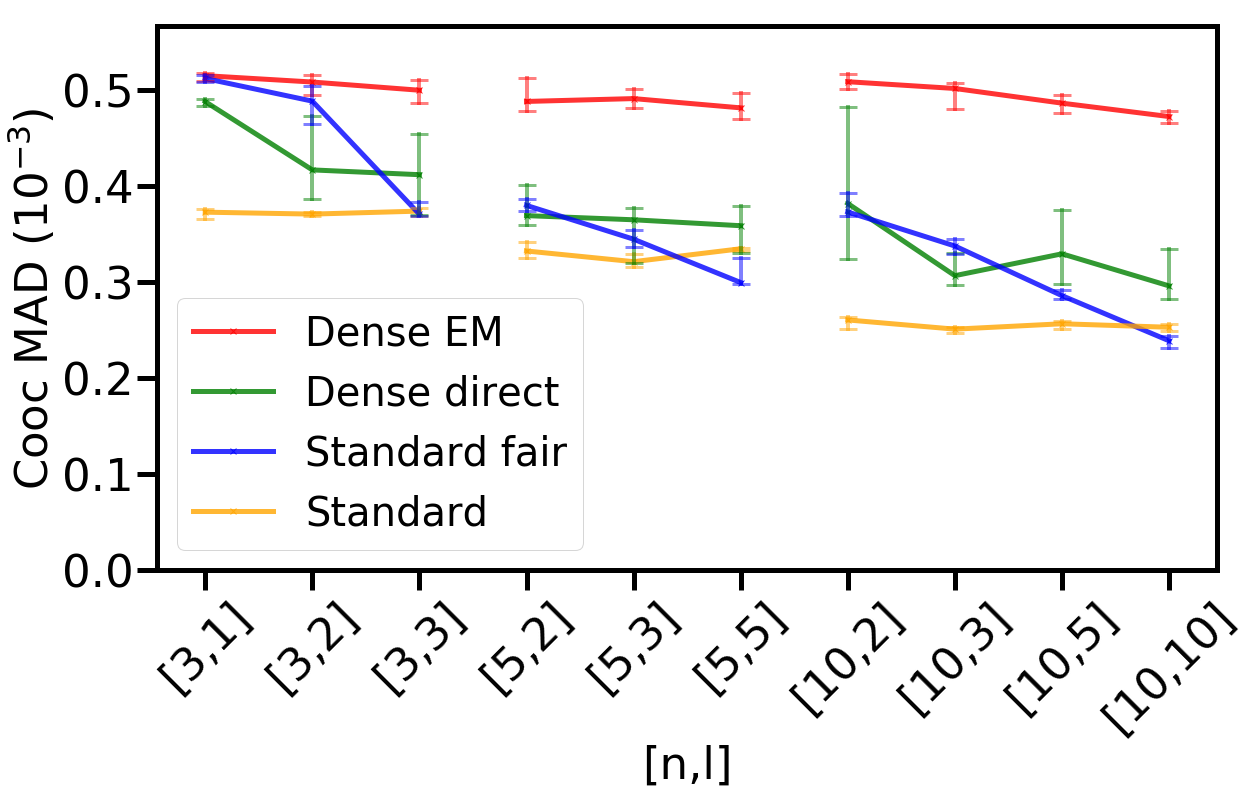}
   \end{subfigure}
    \begin{subfigure}{.5\textwidth}
    \centering
    \includegraphics[height=0.198\textheight]{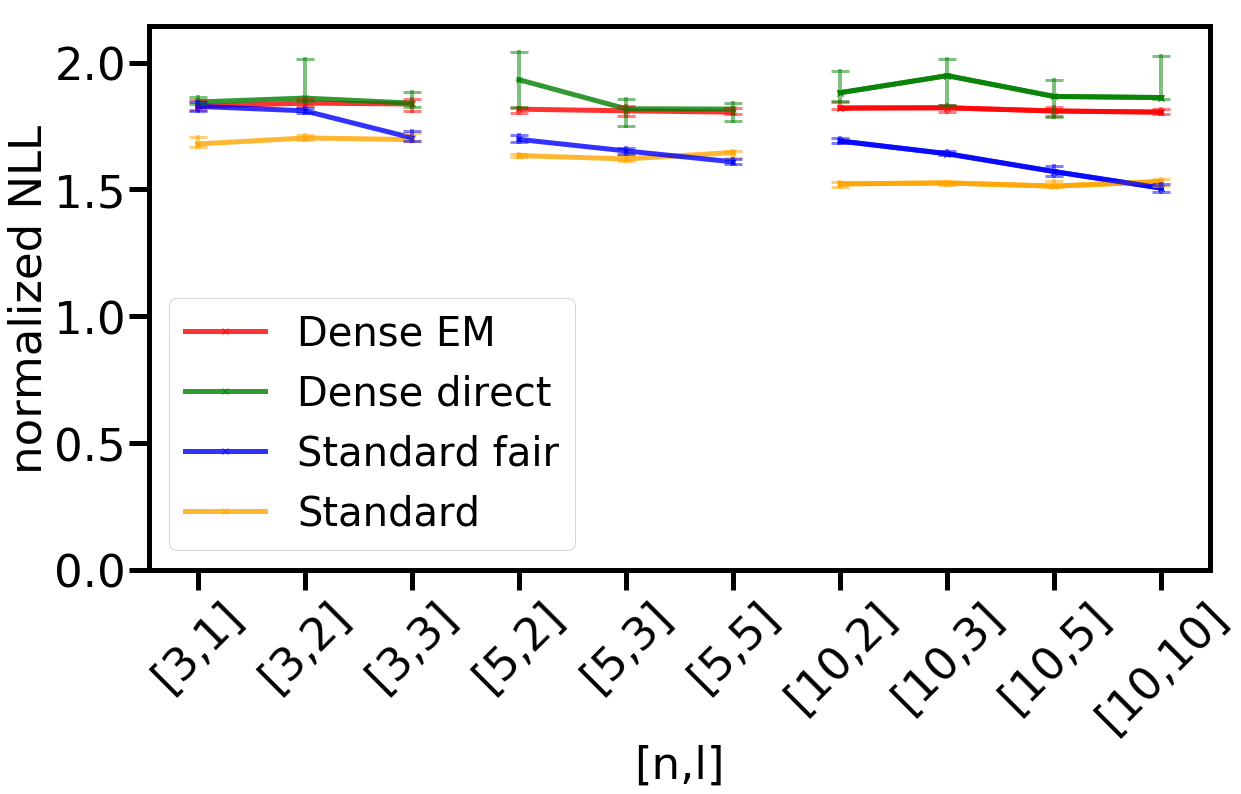}
    \end{subfigure}
    \caption{Co-occurrence mean absolute deviation (left) and normalized negative log-likelihood (right) of the models \denseem~, \densedirect~, \standfair~, \stand~ on part-of-speech tag sequences (Medpost) evaluated for multiple combinations of $n$ and $l$.}
    \label{fig:penntree}
\end{figure}

\paragraph{Part-of-speech sequences} 
The MedPost dataset \cite{medpost} consists of $5{,}700$ sentences. Each sentence consists of words that are tagged with one of $60$ part-of-speech items. Sequences of part-of-speech tags are considered such that each sequence corresponds to one sentence. We apply preprocessing similar to the protein dataset (more details in appendix \ref{appendix:datapreproc}).
The sequences are randomly shuffled and train and test data is split 50:50 for each experiment. Figure \ref{fig:penntree} left shows performance of our models in terms of co-occurrence MADs.
We see that the performance of the standard HMM models as well as \densedirect~ increases with increasing $n$. The number of hidden states seems to be a major driver of performance. Accordingly, \standfair improves with growing $n_{\rm fair}(l) \propto l$. Plus, we have $n_{\rm fair} > n$ for $l \approx n$ which fully explains why \standfair~ is the best performing model in these cases. Overall, \densedirect~ performs competitive to \standfair, especially for the practically more relevant cases with $l < n$. Similar to the other datasets, \denseem~ performs worse than the other models and is barely affected by increasing $n$. Both DenseHMM models have increasing performance for increasing $l$. Normalized NLLs (Figure \ref{fig:penntree} right) are best for the standard HMM models. Both DenseHMM models achieve similar normalized NLLs, which are slightly worse than the ones achieved by the standard HMM models.

\section{Discussion}\label{sec:discussion}

We learn hidden Markov models by learning dense, real-valued vector representations for its hidden and observable states. The involved softmax non-linearity enables to learn high-rank transition matrices, \ie prevents that the matrix ranks are immediately determined by the chosen (in most cases) low-dimensional representation length. We successfully optimize our models in two different ways and find direct co-occurrence optimization to yield competitive results compared to the standard HMM. This optimization technique requires only one gradient descent procedure and no iterative multi-step schemes. It is highly scalable with training data size and also with model size - as it is implemented in a modern deep-learning framework. The optimization is stable and does neither require fine-tuning of learning rate nor of the representation initializations. We release our full tensorflow code to foster active use of DenseHMM in the community.

We leave it to future work to adapt DenseHMMs to HMMs with continuous emissions and study variants of DenseHMM with fewer kinds of learnable representations. First experiments with DenseHMMs that learn only $\bm{Z}$ and $\bm{V}$ lead to almost comparable model quality. From a practitioner's viewpoint, it is worth to investigate how DenseHMM and the learned representations perform on downstream tasks. Using the MedPost dataset, one could consider part-of-speech labeling of word sequences via the Viterbi algorithm after identifying the hidden states of the model with a pre-defined set of ground truth tags. For the protein dataset, a comparison with LSTM-based and BERT embeddings \cite{BeplerB19,min2019pre} could help to understand similarities and differences resulting from modern representation learning techniques.
An analysis of geometrical properties of the learned representations seems promising for systems with $l \gg 1$ as $\exp(l)$ many vectors can be almost-orthogonal in $\mathbb{R}^l$. The $\bm{V}$ representations are a natural choice for such a study as they directly correspond to observation items. The integration of Bayesian optimization techniques like MCMC and VI with DenseHMM is another research avenue.

\section*{Acknowledgement}
The research of J.\ Sicking and T.\ Wirtz was funded by the Fraunhofer Center for Machine Learning within the Fraunhofer Cluster for Cognitive Internet Technologies. The work of M.\ Pintz and M.\ Akila was funded by the German Federal Ministry of Education and Research, ML2R - no. 01S18038B. The authors thank Jasmin Brandt for fruitful discussions.

\bibliography{ref.bib}
\bibliographystyle{ieeetr}
\newpage

\appendix

\section{Full Lagrangians of standard HMM and DenseHMM}\label{appendix:full_lagrangians}
The full Lagrangian of the standard HMM model in the M-step reads

\begin{flalign*}
    \bar{\mathcal{L}} =&\ \ \ \bar{\mathcal{L}}_1 + \bar{\mathcal{L}}_2 + \bar{\mathcal{L}}_3&\\
    =&\sum\limits_{i, j \in [n]} \sum_{t=2}^T \gamma_t(s_i, s_j) \log a_{ij} + \sum_{i \in [n]} \varphi_i \Big(1 - \sum_{j \in [n]} a_{ij}\Big)&\\
    &+\sum_{i \in [n]} \sum_{t=1}^T \gamma_t(s_i) \log b_{i, j_{o_t}} + \sum_{i \in [n]} \varepsilon_i \Big(1 - \sum_{j\in [m]} b_{ij}\Big)& \\
    &+\sum_{i \in [n]} \gamma_1(s_i) \log \pi_{i} + \bar{\varphi} \Big(1 - \sum_{i \in [n]} \pi_i\Big),&
\end{flalign*}
where $j_{o_t}$ describes the index of the observation observed at time $t$ and $\bar{\varphi}, \varepsilon_i$ are Lagrange multipliers. Applying the transformations $\mathbf{A} = \mathbf{A}(\mathbf{U}$,\,$\mathbf{Z})$, $\mathbf{B}=\mathbf{B}(\mathbf{V}$,$\mathbf{W})$ and $\mathbf{\pi}=\mathbf{\pi}(\mathbf{U},\mathbf{z}_{\rm start})$ yields the full Lagrangian of the DenseHMM:
 
\begin{flalign*}
    \bar{\mathcal{L}}^{\rm{dense}} =&\ \ \ \ \ \bar{\mathcal{L}}^{\rm dense}_1 + \bar{\mathcal{L}}^{\rm dense}_2 + \bar{\mathcal{L}}^{\rm dense}_3&\\
    =&\sum_{i, j\in [n]} \sum_{t=2}^T \gamma_t(s_i, s_j)\,\vec{u}_{j} \cdot \vec{z}_{i} 
    - \sum_{i, j \in [n]} \sum_{t=2}^T \gamma_t(s_i, s_j) \log \sum_{ k \in [n]} \exp(\vec{u}_k \cdot \vec{z}_{i})&\\
    &+ \sum_{i \in [n]} \sum_{t=1}^T \gamma_t(s_i)\,\vec{v}_{j_{o_t}} \cdot \vec{w}_{i}
    - \sum_{i \in [n]} \sum_{t=1}^T \gamma_t(s_i) \log \sum_{ j \in [m]} \exp(\vec{v}_j \cdot \vec{w}_{i})  &\\
    &+ \sum_{i\in [n]} \gamma_1(s_i)\,\vec{u}_{i} \cdot \vec{z}_{start} 
    - \sum_{i \in [n]} \gamma_1(s_i)  \log \sum_{ j \in [n]} \exp(\vec{u}_j \cdot \vec{z}_{start}).&
\end{flalign*}

\section{Non-linear A-matrix factorization}\label{appendix:non_lin_factorization}
All matrix sizes $n$ and representation lengths $l$ that contribute to the visualized $l/n$ ratios in Figure \ref{fig:nonlinear_matrix_approx} are shown in Table \ref{expd}.

\section{Implementation details and data preprocessing} \label{appendix:datapreproc}

\paragraph{Implementation details} The backbone of our implementation is the library \verb|hmmlearn|\footnote{\url{https://github.com/hmmlearn/hmmlearn}} that provides functions to optimize and score HMMs. The  optimization schemes for the DenseHMM models \denseem~ and \densedirect~ are implemented in tensorflow~\cite{abadi2016tensorflow}. Both models use \verb|tf.train.AdamOptimizer| with a fixed learning rate for optimization. At this point we note that experiments done with other optimizers such that \verb|tf.train.GradientDescentOptimizer| lead to similar results in the evaluation. The representations are initialized using a standard isotropic Gaussian distribution. ${\rm NLL}$ values are normalized by the number of test sequences and by the maximum test sequence length. 

\paragraph{Hardware used} All experiments are conducted on a \verb|Intel(R) Xeon(R) Silver 4116 CPU| with \verb|2.10GHz| and a \verb|NVidia Tesla V100|.

\paragraph{Protein dataset preprocessing} The first $1{,}024$ sequences of the RCSB PDB dataset have $22$ unique symbols. We cut each sequence after a length of $512$. Note that less than $4.9\%$ of the $1{,}024$ sequences exceed that length. Additionally, we collect the symbols of lowest frequency that together make up less than $0.2\%$ of all symbols in the sequences and map them onto one residual symbol. This reduces the number of unique symbols in the sequences from $22$ to $19$.

\paragraph{Part-of-speech sequences preprocessing} We take ${1},{000}$ sequences from the Medpost dataset (from \verb|tag_mb.ioc|) and cut them after a length of $40$ which affects less than $15\%$ of all sequences. We also collect the tags of lowest frequency that together make up less than $1\%$ of all tags in the sequences and map them onto one residual tag. This reduces the number of tag items from $60$ to $\mathit{42}$. 

\paragraph{Calculation of $\bm{\Omega}^{\rm gt}$ and $\bm{\Omega}^{\rm model} $}  The co-occurrence matrices $\bm{\Omega}^{\rm model}$ and $\bm{\Omega}^{\rm{gt}}$ used to calculate the co-occurrence MADs in section \ref{sec:emp_evaluation} are estimated by counting subsequent pairs of observation symbols $(o_i(t), o_j(t+1)) \in O^2$. For real-world data, $\bm{\Omega}^{\rm{gt}}$ is estimated based on the test data ground truth sequences. Equally long sequences sampled from the trained model are used to estimate $\bm{\Omega}^{\rm model}$. In case of synthetic data, $\bm{\Omega}^{\rm{gt}}$ is calculated analytically (eq. \ref{eq:coocs}) instead.

\begin{table}[h] 
\centering
\caption{Approximation errors (median with 25/75 percentile) of normAbsLin-based and softmax-based matrix factorizations for different matrix sizes $n$ and representation \mbox{lengths $l$.}}\label{expd}
\vspace{0.8em}
\begin{tabular}{l l l l}
\toprule
  {\bfseries n} & {\bfseries l} & \multicolumn{2}{ c }{{\bfseries median (25/75 percentile) of \textbf{loss}$(\bf{\tilde{A}},\bf{A_{gt}})$}} \\
    \cline{3-4}
    \addlinespace[0.5em]
  & & $\tilde{A} = \rm{normAbsLin}(\mathbf{UZ})$ & $\tilde{A} = {\rm softmax}(\mathbf{UZ})$ \\
\bottomrule
  \addlinespace[0.5em]
  3 \hspace{0cm}& 1\hspace{1cm} & 0.678 (0.652/0.696) & { \textbf{0.048} (0.004/0.110)} \\
  3 & 2 & 0.162 (0.002/0.270) & { \textbf{0.001} (0.000/0.001)} \\
  3 & 3 & \textbf{0.001} (0.001/0.001) & { \textbf{0.001} (0.000/0.001)} \\
  3 & 5 & \textbf{0.001} (0.001/0.001) & { \textbf{0.001} (0.001/0.001)} \\
  \addlinespace[0.4em]
  5 & 1 & 0.769 (0.745/0.827) & { \textbf{0.453} (0.321/0.505)} \\
  5 & 3 & 0.346 (0.093/0.396) & { \textbf{0.001} (0.001/0.003)} \\
  5 & 5 & \textbf{0.001} (0.001/0.002) & { \textbf{0.001} (0.001/0.001)} \\
  5 & 10 & \textbf{0.001} (0.001/0.002) & { 0.002 (0.001/0.003)} \\
  \addlinespace[0.4em]
  10 & 1 & 0.862 (0.851/0.868) & { \textbf{0.616} (0.581/0.645)} \\
  10 & 5 & 0.310 (0.235/0.345) & { \textbf{0.012} (0.005/0.028)} \\
  10 & 10 & \textbf{0.002} (0.002/0.002) & { 0.003 (0.002/0.005)} \\
  10 & 15 & \textbf{0.002} (0.002/0.002) & { 0.003 (0.003/0.043)} \\
\bottomrule
  \addlinespace[0.8em]
\end{tabular}
\end{table}

\end{document}